\documentclass[10pt,twocolumn,letterpaper]{article}

\usepackage[pagenumbers]{wacv}

\usepackage{times}
\usepackage{epsfig}
\usepackage{graphicx}
\usepackage{amsmath}
\usepackage{amssymb}
\usepackage{booktabs}

\usepackage{bm}
\usepackage{multirow}
\usepackage{tabularx}
\usepackage{xcolor}
\usepackage{colortbl}
\newcolumntype{P}[1]{>{\centering\arraybackslash}p{#1}}
\newcolumntype{M}[1]{>{\centering\arraybackslash}m{#1}}
\usepackage{diagbox}\newcommand\modelname{\texttt{Debiasify}}
\newcommand{\mycustomfont}{\fontsize{6.9pt}{8pt}\selectfont}

\usepackage{todonotes}
\usepackage{tikz}
\newcommand{\tikzcircle}[2][red,fill=red]{\tikz[baseline=-0.5ex]\draw[#1,radius=#2] (0,0) circle ;}%
\usepackage{comment}

\usepackage[pagebackref,breaklinks,colorlinks]{hyperref}

\usepackage[capitalize]{cleveref}
\crefname{section}{Sec.}{Secs.}
\Crefname{section}{Section}{Sections}
\Crefname{table}{Table}{Tables}
\crefname{table}{Tab.}{Tabs.}

\def\wacvPaperID{2150} % *** Enter the WACV Paper ID here
\def\confName{WACV}
\def\confYear{2025}

\begin{document}

\title{\modelname{}: Self-Distillation for Unsupervised Bias Mitigation}

\author{
    Nourhan Bayasi\textsuperscript{1}\thanks{Equal contribution.} \quad
    Jamil Fayyad\textsuperscript{2}\footnotemark[1] \quad
    Ghassan Hamarneh\textsuperscript{3} \quad
    Rafeef Garbi\textsuperscript{1} \quad
    Homayoun Najjaran\textsuperscript{2} \\
    \textsuperscript{1}University of British Columbia \quad
    \textsuperscript{2}University of Victoria \quad
    \textsuperscript{3}Simon Fraser University \\
    \small \texttt{nourhanb,rafeef@ece.ubc.ca} \quad \texttt{jfayyad,najjaran@uvic.ca} \quad \texttt{hamarneh@sfu.ca}
}
\maketitle

%%%%%%%%% ABSTRACT
\begin{abstract}
Simplicity bias is a critical challenge in neural networks since it often leads to favoring simpler solutions and learning unintended decision rules captured by spurious correlations, causing models to be biased and diminishing their generalizability. While existing solutions rely on human supervision, obtaining annotations of the different bias attributes is often impractical. To tackle this, we present \modelname{}, a novel self-distillation approach that works without any prior information about the nature of biases. Our method leverages a new distillation loss to distill knowledge within a network; from a deep layer where complex, highly-predictive features reside, to a shallow layer where simpler yet attribute-conditioned features are found in an unsupervised manner. In this way, \modelname{} learns robust, debiased representations that generalize well across various biases and datasets,  enhancing worst-group performance and overall accuracy. Extensive experiments on computer vision and medical imaging benchmarks show the efficacy of our method, significantly outperforming the previous unsupervised debiasing methods (e.g., a 10.13\% improvement in worst-group accuracy on Wavy Hair classification in CelebA) while achieving comparable or superior performance to supervised methods. Our code is publicly available at the following link:\href{https://github.com/nourhanb/Debiasify} {Debiasify}.
\end{abstract}

%%%%%%%%% BODY TEXT
\section{Introduction}
Deep neural networks have emerged as a fundamental technology in numerous applications that profoundly impact various aspects of society, such as facial recognition~\cite{jung2021fair}, AI-enabled recruitment~\cite{kochling2020discriminated}, and healthcare diagnostics~\cite{du2022fairdisco,bayasi2024biaspruner}. Given the significant societal implications of these algorithms, it is increasingly crucial to ensure their resilience against \textit{simplicity bias}~\cite{shah2020pitfalls,pezeshki2021gradient}; in other words, these networks' learning process should not prioritize weak predictive features over complex features that underpin the actual mechanisms of the task of interest. For instance, on the CelebA dataset~\cite{liu2018large},  which is a real-world dataset where different attributes are strongly correlated, networks tend to classify hair color based on gender, frequently associating \texttt{Blond Hair} with \texttt{Female}. Such an unintended rule  performs adequately across the majority of training instances but leads to unforeseen extreme errors in minority examples that 
lack the spurious correlation, thereby hindering the model's ability to adapt to new testing scenarios that exhibit changes in data distributions.

Effective ways for network debiasing include upweighting or upsampling of examples that lack spurious correlations~\cite{niu2022roadblocks}, data augmentation~\cite{duboudin2022look}, adversarial learning~\cite{tiwari2023overcoming,du2022fairdisco}, robust learning~\cite{pezeshki2021gradient}, and architecture optimization~\cite{bai2021ood}. Nevertheless, most of these efforts rely on explicit bias attribute labels in their debiasing recipes. This compromises their practicality,  as identifying and manually labeling the types of biases,  to determine which attributes involve spurious correlations without a thorough analysis of the model and dataset, present significant challenges. Only recently, the focus has been shifted towards debiasing without the bias attribute labels. This is usually achieved by identifying the minority group within each class -- flagged based on indicators such as misclassification~\cite{liu2021just}, high loss~\cite{nam2020learning}, or sensitive representations~\cite{creager2021environment}, and subsequently upweighting/upsampling them during training.  

Despite being promising, these methods have two major drawbacks. First, they are heavily dependent on {hyperparameter tuning using  bias attribute information in the validation set, which might not be accessible for datasets in the real world}~\cite{kirichenko2022last}.  Second, they are designed to address only a single bias attribute within a class, neglecting the potential existence of multiple bias sources  within the same class; e.g., skin tone, gender, image background.  
\begin{figure}
    \centering
    \includegraphics[width=1\linewidth]{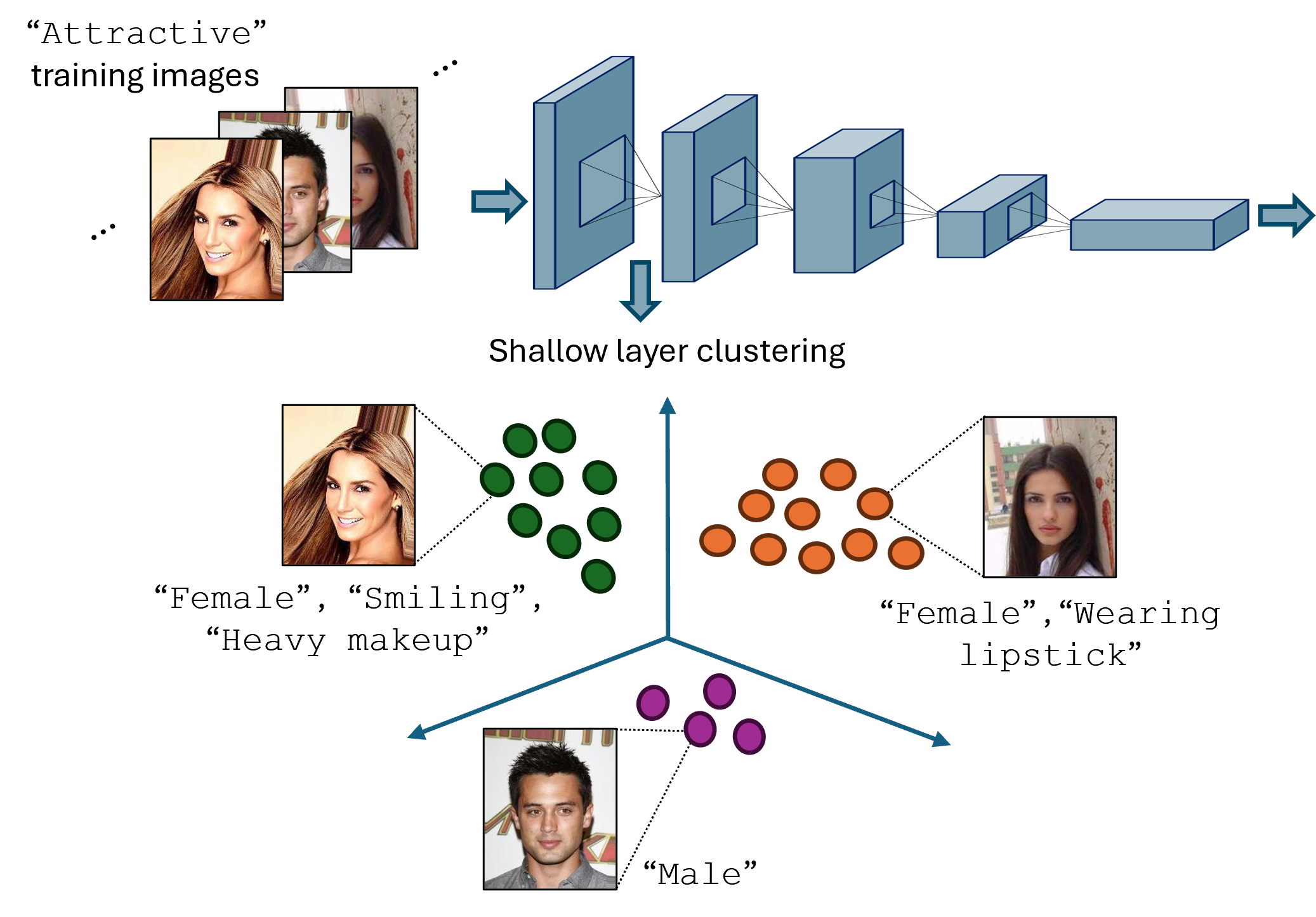}
    \caption{\modelname{} leverages clustering in the feature space of a shallow layer in the network to identify attribute-conditioned groups (3 groups shown) for each class (e.g., \texttt{Attractive}), where images in each group are clustered based on common, non-target bias attributes (e.g., \texttt{Female}, \texttt{Smiling}, etc.)}

    \label{fig:overview}
\end{figure}

To overcome the aforementioned problems, we propose \modelname{}, a simple yet effective unsupervised debiasing technique via feature clustering and self-distillation. Rooted in the observations that images sharing the same label for certain bias attribute(s) (other than the target attribute) tend to have similar representations in the feature space, particularly in the shallow layers of the neural network stack~\cite{dagaev2023too,tiwari2023overcoming}, we propose clustering the shallow layer features to identify attribute-conditioned groups for each class, as depicted in Figure~\ref{fig:overview}. To exploit these groups for learning debiased representations, we introduce a novel self-distillation loss that encourages their distributions to converge while simultaneously aligning them with the distribution of their highly-predictive class features in the deep layer. 
In summary, our contributions are:
\begin{itemize}
    \item {We introduce \modelname{}, a new method for unsupervised  bias mitigation  through a self, deep-to-shallow, distillation technique. 
    \item We propose a hybrid loss that maintains high classification performance while effectively debiasing representations by minimizing the distance between class-specific, bias attribute-conditioned groups in the shallow layers and their corresponding class attribute-agnostic distributions in the deep layers.}
    \item We conduct experiments and ablation studies on CelebA, Waterbirds, and Fitzpatrick, benchmarking against bias-unsupervised methods, including previous SOTA: CFix~\cite{capitani2024clusterfix}, and the uper-bound supervised method:  GDRO~\cite{sagawa2019distributionally}. Our results highlight \modelname{}'s superior performance, especially in worst-group accuracy.

\end{itemize}

\section{Related Work}
\subsection{Simplicity Bias in Neural Networks}
Neural networks have been found to be prone to simplicity bias~\cite{shah2020pitfalls,berchenko2024simplicity}. That is, they tend to learn the simplest features to solve a task, even in the presence of other, more robust but more complex features. This bias towards simpler features can lead to models lacking robustness against shifts that do not adhere to the simplistic characteristics captured by the learned features. Extensive efforts have been made to address the simplicity bias problem, categorized broadly into three approaches based on the stage of intervention during the modeling process. Pre-processing techniques aim to modify the training data in order to  reduce the correlations between bias and target attributes~\cite{biswas2021fair,madras2018learning}; in-processing techniques modify the learning algorithms to eliminate bias during the model training process~\cite{du2022fairdisco,chuang2021fair}; and post-processing techniques treat the learned model as a black-box and try to mitigate bias by leveraging the predictions~\cite{petersen2021post,xian2023fair}. Nevertheless, most of these techniques have limitations in real-world scenarios since they rely on access to bias attribute annotations in the training or validation sets for effective bias mitigation. 
\begin{figure*}
    \centering
    \includegraphics[width=1\linewidth]{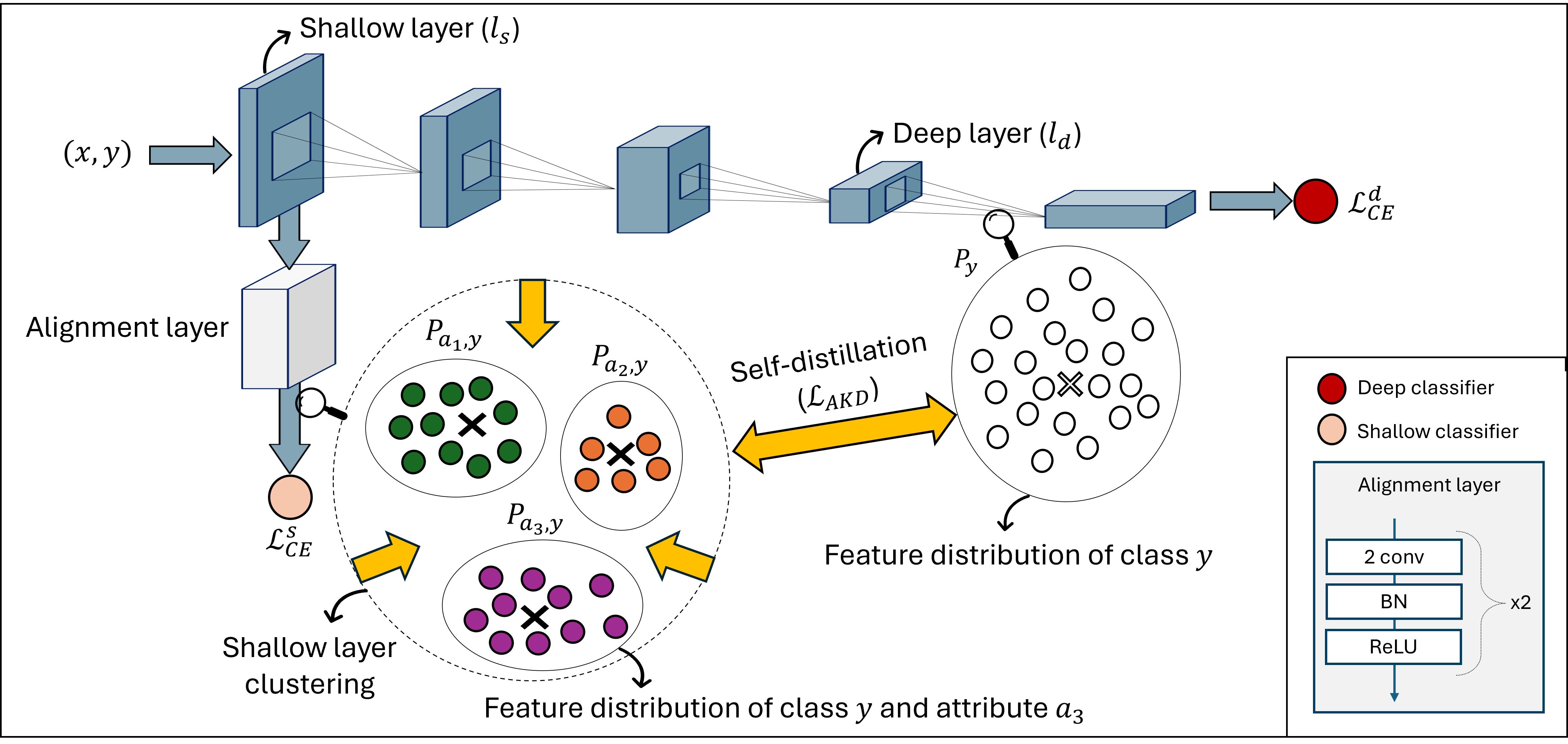}
    \caption{\modelname{} identifies attribute-conditioned groups \( P_{a_{k},y} \) (represented by \tikzcircle[fill=teal]{2pt} \tikzcircle[fill=orange]{2pt} \tikzcircle[fill=violet]{2pt} for all $a_k$ attributes, $k$ = 1, 2, 3) found through clustering in  feature space of a shallow network layer. The goal is to bring their distributions closer to each other while aligning them with their class distribution \( P_{y} \) (represented by \tikzcircle[fill=white]{2pt}) in the deep layer using a novel self-distillation loss $\mathcal{L}_{AKD}$ (yellow arrows). }
    \label{fig:main_fig}
\end{figure*}
\subsection{Bias Mitigation without Supervision}
Recently, efforts have been directed towards mitigating bias in the absence of explicit bias attributes, primarily through in-processing techniques~\cite{zhang2024discover,bayasi2024biaspruner,nahon2023mining,wu2023discover,asgari2022masktune,tiwari2023overcoming}. One such technique involves identifying minority samples as those misclassified by an initial network and then reweighting them. Nam et al.~\cite{nam2020learning} achieve this by training an additional biased model, where images that are not easily trained by the biased model are considered minority. Liu et al.~\cite{liu2021just} define minority samples as those misclassified by a model trained using empirical risk minimization and prioritize them during the training of a debiased model. Another technique synthesizes images having similar characteristics to the minority group and employs them to train a debiased model. Kim et al.~\cite{kim2021biaswap} synthesize images without bias attributes by leveraging an image-to-image translation model. Lee et al.~\cite{lee2021learning} and Hwang
et al.~\cite{hwang2022selecmix} augment minority samples in the feature space by employing disentangled representations and mixup, respectively. 

Our work aligns closely with a third technique, which involves acquiring bias pseudo-labels through the unsupervised learning technique of clustering in the feature space of the network. Examples in this category include BPA~\cite{seo2022unsupervised}, which proposes a cluster-wise reweighting scheme, leveraging pseudo-attribute information from feature clustering results; CFix~\cite{capitani2024clusterfix}, which uses cluster error (i.e., the discrepancy in correctly classifying examples within clusters) to identify examples potentially influenced by network inductive bias, subsequently upweighting them to enhance worst-group performance; and George~\cite{sohoni2020no}, which approximates bias attributes with cluster assignment and weights the objective function to maximize the worst-group accuracy. However, these clustering-based methods have a key limitation: they rely solely on reweighting of images within their respective clusters, which can be problematic given that formation of clusters (their size and shape) is sensitive to outliers and noisy images that are commonly encountered in real-world datasets.{ In contrast, our \modelname{} takes a fundamentally different approach by naturally examining cluster distributions and promoting their alignment through a novel self-distillation loss, without any reweighting.}

\section{Methodology}
Figure~\ref{fig:main_fig} presents an overview of our proposed self-distillation method for unsupervised bias mitigation, termed \modelname{}. Our self-distillation loss directs a shallow network layer to learn more predictive features instead of simpler ones that might be correlated with unwanted characteristics to improve performance. Assuming bias information is unknown, \modelname{} first performs clustering in the feature space of a shallow layer to identify class-wise attribute-conditioned groups, which are then guided to get closer to each other while simultaneously mirroring their class, attribute-agnostic counterparts in the deep layer. Observations and details are given next. 

\subsection{Observations} \label{sec:prel_results}
\noindent \textbf{Objective. } Let $\mathbf{x}$ be an image associated with a set of possible bias attributes $a \in {\mathcal{A}}$. The primary objective of our \modelname{} is to predict a target attribute $y$ by estimating the ground truth relationship $p\left(y \mid \mathbf{x}\right)$, while mitigating any undesired correlations with other bias attributes; i.e., ensuring that $p\left(y \mid \mathbf{x}\right)=p\left(y \mid \mathbf{x}, a\right), \forall ~ a \in {\mathcal{A}}$. During the training phase of \modelname{}, information regarding bias set ${\mathcal{A}}$ is neither available nor provided.

\begin{figure}
    \centering
    \includegraphics[width=1\linewidth]{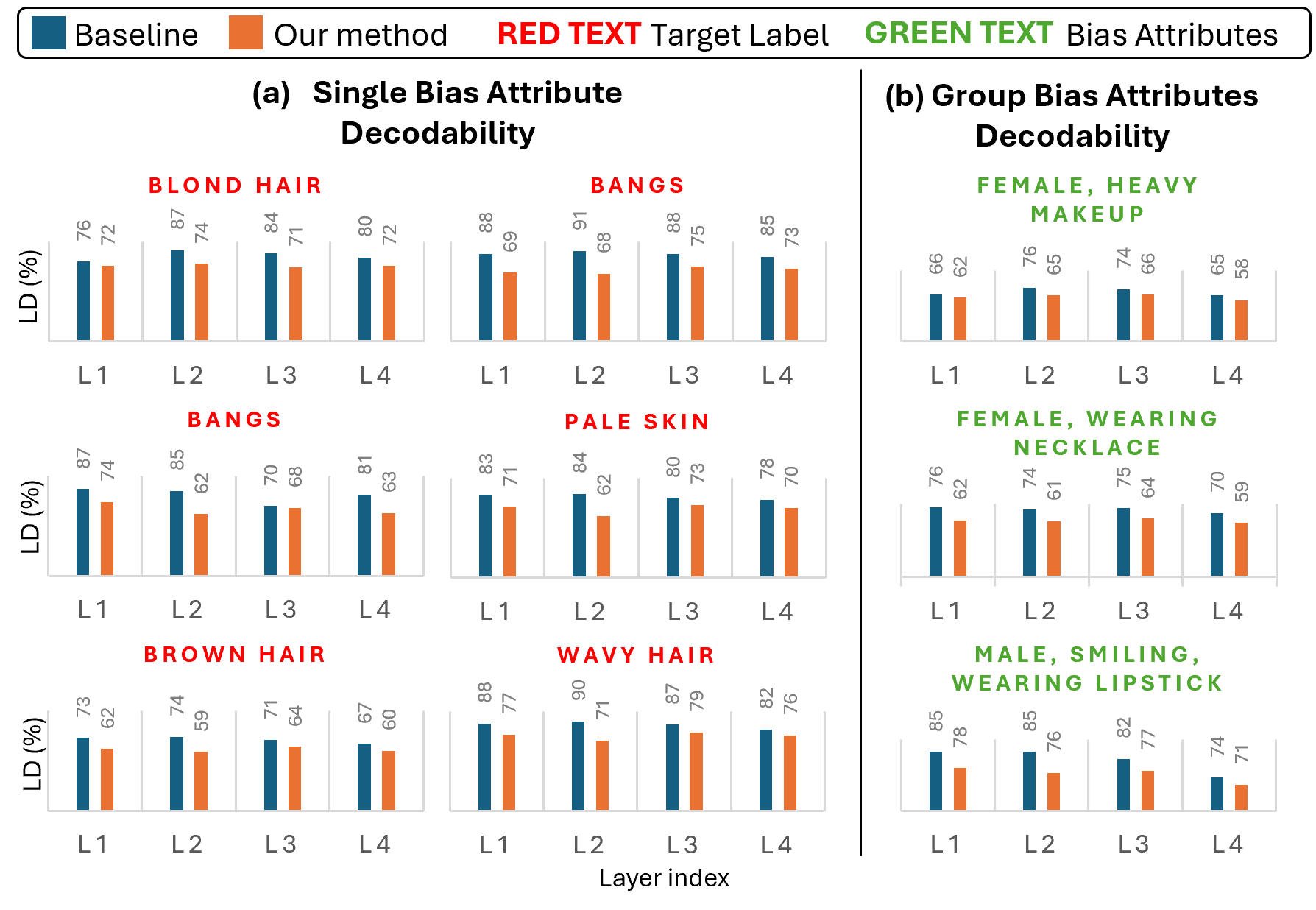}
    \caption{Results of Linear Decodability (LD): Panel (a) compares LD of the \texttt{Male} bias attribute from a frozen baseline network (blue) and our method (orange), both pretrained on different target attributes. Panel (b) shows LD of multiple bias attributes from networks pretrained on \texttt{Blond Hair}, comparing the baseline (blue) and our method (orange).}
    \label{fig:linear_dec}
\end{figure}

\noindent \textbf{Preliminary Study.} \label{prelim_results}
Previous works~\cite{seo2022unsupervised,capitani2024clusterfix} have used the CelebA dataset to analyze the feature semantics over the target and bias attributes. The studies revealed that images from certain groups, defined by a combination of target and bias attribute values, are clustered in the feature space, even without using the bias information during training. Expanding on their analysis, we conduct the following experiments to investigate the layer where the clustering could be more pronounced. We begin by training a ResNet18 model for 50 epochs to classify a target attribute (e.g., \texttt{Blond Hair}). Next, we assess feature decodability at different network depths to evaluate how well the bias attribute \texttt{Male} can be decoded, keeping the network parameters frozen. A decoder, consisting of a single linear layer and softmax, is trained using activations from a specific layer and an unbiased validation set labeled with the bias attribute.  Figure~\ref{fig:linear_dec} -- (a, blue bars) shows the decodability of \texttt{Male} across layers. We observe that bias decodability generally decreases with network depth, indicating that shallower layers are more effective at detecting bias. In more complex scenarios, where the bias involves combinations of attributes, a similar trend is observed (Figure~\ref{fig:linear_dec} -- (b, blue bars)). These findings align with previous studies that identified bias detection in shallow layers~\cite{tiwari2024using,tiwari2023overcoming,dagaev2023too} and were theoretically validated~\cite{hermann2020shapes}.

\subsection{Formulation} \label{sec:framework}
\noindent \textbf{Preliminaries.} We aim to train a network $f$ to enhance the classification  performance and generalization by learning debiased representations. We denote the shallow and deep layers of $f$ as $l_s$ and $l_d$, respectively. To align the feature distributions across $l_s$ and $l_d$, a simple auxiliary branch is introduced after $l_s$, comprising an alignment layer and a classifier, as illustrated in Figure~\ref{fig:main_fig}. The alignment layer ensures that the feature dimensionality in $l_s$ matches that of $l_d$. Additionally, we denote the predicted labels of sample $\mathbf{x}$ by the shallow and deep classifiers as $c_s$ and $c_d$, respectively.

\noindent \textbf{Generating Attribute-conditioned Groups.} We begin by training $f$ for a few epochs using an averaged cross entropy loss $\mathcal{L}_{A C E}$ to predict the ground truth $y$ for each sample $\mathbf{x}$, as follows;
\begin{equation}
\mathcal{L}_{A C E}  = \frac{1}{2} \sum_y \left[ \mathcal{L}^s_{CE}\left(c_s, y\right) + \mathcal{L}^d_{CE}\left(c_d, y\right) \right] . 
\label{eq:ace}
\end{equation}
Once the network is trained and potentially learned the bias, we cluster the feature embeddings from $l_s$ to generate $\mathcal{K}$ attribute-conditioned groups for each class, given our empirical observations in Sec~\ref{sec:prel_results} and previous findings confirming that clustering effectively groups images sharing common non-target attributes. In our implementations, we opt to use K-means~\cite{lloyd1982least} with adaptive $\mathcal{K}$ value to mitigate the risk of failing to capture smaller clusters if $\mathcal{K}$ is small or the emergence of unwanted clusters if $\mathcal{K}$ is large. We determine the  value of $\mathcal{K}$ as  the smallest integer such that the mean within-cluster variance of the features is lower than a pre-defined upper bound $\gamma$.  As a result, we obtain a set of clusters that comprehensively cover each class,  where images within each cluster share common attributes other than the target attribute.

\noindent \textbf{Learning Debiased Representations. } Next, we aim to learn debiased representations by minimizing the discrepancy between the class-wise feature distribution \( P_y \) extracted from \( l_d \), and the distribution of each attribute-conditioned cluster \( P_{a_k,y} \) derived from \( l_s \) and associated with the same class. Note that \( k \) ranges from 1 to \( \mathcal{K} \). Specifically, we define our attribute-based knowledge distillation loss as follows; 
\begin{equation}
\mathcal{L}_{AKD} = \sum_y \sum_k \mathrm{D}^2\left(P_y, P_{a_k,y}\right),
\label{eq:akd}
\end{equation}
\noindent where $\mathrm{D}$ is a distance metric between two distributions. In our experiments, we use the Maximum Mean Discrepancy (MMD)~\cite{gretton2012kernel}, a powerful method for comparing distributions without relying on specific assumptions about their shapes. MMD works by evaluating the difference in average values across functions in a special space called a Reproducing Kernel Hilbert Space (RKHS). To facilitate this comparison, it employs a kernel, such as the Gaussian Radial Basis Function (RBF) in our context, to convert the distributions into the RKHS, thereby enabling easier analysis. We experiment with different distance metrics in Sec.~\ref{sec:ablations}.
By optimizing \(\mathcal{L}_{AKD}\), i.e., $
\mathrm{D}^2(P_y, P_{a_k,y}) \rightarrow 0 \implies P_{a_k,y} \approx P_y $, the group features are encouraged to mirror their respective class distribution, allowing the model to learn representations that capture the unique characteristics of each class while washing out the attribute-specific features that could lead to biased learning. The final objective for \modelname{}'s training is given as follows;
\begin{equation}
    \mathcal{L}_{hybrid} = \mathcal{L}_{A C E} +  \alpha\mathcal{L}_{A K D}  + \mathcal{L}_{K L} ~ , \
    \label{eq:hybrid}
\end{equation}
\noindent where $\mathcal{L}_{K L} = \text{KL} (c_s, c_d)$ is the Kullback-Leibler divergence between the shallow and deep classifier logits, and $\alpha$ is a hyperparameter that controls the balance between classification accuracy and knowledge transfer.

\section{Experiments}
We conduct experiments to assess the performance of \modelname{} across various benchmarks and compare it with state-of-the-art (SOTA) methods for bias mitigation. To ensure a fair comparison with other clustering-based debiasing methods, we adopt the experimental settings in the previous SOTA, CFix~\cite{capitani2024clusterfix}, when applicable.

\subsection{Evaluation Benchmarks}
We evaluate \modelname{} using three datasets: CelebA~\cite{liu2015deep}, Waterbirds~\cite{sagawa2019distributionally} and Fitzpatrick~\cite{groh2021evaluating}. 

\noindent \textbf{CelebA} is a real-world dataset for face attribute recognition containing 202,599 celebrity face images, each annotated with 40 binary attributes. Following other works~\cite{capitani2024clusterfix,seo2022unsupervised}, we designate \texttt{Male} as the bias attribute, and we diversify the target label by selecting other attributes that exhibit the strongest correlation with the bias. 

\noindent \textbf{Waterbirds} is a synthesized dataset created by combining two other datasets to establish a strong correlation between bird types ({waterbird, landbird}) and their backgrounds ({water, land}). It consists of 4,795 training examples; 95\% of them having matching bird types and backgrounds; e.g., waterbirds with water background, while the other do not.

\noindent \textbf{Fitzpatrick} is a well-known medical dataset for skin lesion analysis consisting of 16,012 clinical images with 3 class labels. Each image is annotated with a Fitzpatrick score representing the skin tone, which we designate as the bias attribute. The Fitzpatrick scale consists of six scores, ranging from 1 (the lightest skin tone) to 6 (the darkest skin tone). As in previous work~\cite{wu2022fairprune}, we group skin tones 1 to 3 into a category representing lighter skin tones, and skin tones 4 to 6 into a category representing darker skin tones. The dataset is imbalanced, with significantly more images of lighter skin tones compared to darker skin tones (11,060 vs. 4,952, respectively).

\begin{table*}
\mycustomfont	
\caption{Performance evaluation of \modelname{} against others on CelebA dataset. Cells in  \textcolor{blue!50}{blue} and \textcolor{green!70}{green} represent the best and second-best results, respectively. Improvement gain compared to the second-best method is given between brackets in red.}
\centering
\begin{tabular}{|
>{\centering\arraybackslash}p{32mm}|
>{\centering\arraybackslash}p{14mm}
>{\centering\arraybackslash}p{14mm}
>{\centering\arraybackslash}p{14mm}
>{\centering\arraybackslash}p{14mm}
>{\centering\arraybackslash}p{14mm}
>{\centering\arraybackslash}p{22mm}|
>{\centering\arraybackslash}p{14mm}|} 
\multicolumn{7}{c}{\textbf{(a) Unbiased Accuracy (\%) }} \\ \hline
& \multicolumn{6}{c}{\cellcolor{gray!25}\textbf{Unsupervised}} & {\cellcolor{gray!25}\textbf{Supervised}}  \\ \hline
\textbf{Target} & \textbf{Baseline} & \textbf{LfF} & \textbf{George} &  \textbf{BPA} & \textbf{CFix} & \textbf{Ours} & \textbf{GDRO} \\ \hline
\textbf{\texttt{Double Chin}} & 64.61$\pm$0.82 &68.47$\pm$0.22 &76.23$\pm$0.11& 82.92$\pm$0.54 &\cellcolor{green!10}{85.13$\pm$0.30}& \cellcolor{blue!15}\textbf{86.19$\pm$0.21} \textcolor{red}{\tiny{(+1.06)}}& 83.19$\pm$1.11 \\ 

\textbf{\texttt{Pale Skin}}  &71.50$\pm$1.60& 75.23$\pm$0.74 &78.22$\pm$3.75& 90.06$\pm$0.75 &\cellcolor{green!10}91.17$\pm$0.04&\cellcolor{blue!15}\textbf{92.77$\pm$1.38} \textcolor{red}{\tiny{(+1.60)}}& 90.55$\pm$0.84 \\

\textbf{\texttt{Wearing Necklace}}  &55.04$\pm$0.59& 57.21$\pm$0.76& 58.79$\pm$0.10 &68.96$\pm$0.12 &\cellcolor{green!10}68.99$\pm$1.19 &\cellcolor{blue!15} \textbf{71.14$\pm$0.52} \textcolor{red}{\tiny{(+2.15)}} &62.89$\pm$3.69 \\

\textbf{\texttt{Wearing Hat}}& 93.53$\pm$0.37 &94.81$\pm$0.15 &95.72$\pm$0.71 &96.80$\pm$0.26& \cellcolor{blue!15} \textbf{97.88$\pm$0.09}&\cellcolor{green!10}97.65$\pm$0.37 \textcolor{red}{\tiny{(--0.23)}}& 96.84$\pm$0.46 \\

\textbf{\texttt{Big Lips}} &60.87$\pm$0.58& 62.15$\pm$0.06& 64.99$\pm$0.13& \cellcolor{green!10}{66.50$\pm$0.24}& 65.40$\pm$0.48& \cellcolor{blue!15}{\textbf{67.73$\pm$0.44}} \textcolor{red}{\tiny{(+1.23)}}& 63.70$\pm$0.44 \\

\textbf{\texttt{Bangs}} &89.04$\pm$0.47 &89.04$\pm$0.50& 92.62$\pm$0.12& 93.94$\pm$0.57& \cellcolor{green!10}94.67$\pm$0.16 & \cellcolor{blue!15} \textbf{95.56$\pm$0.93} \textcolor{red}{\tiny{(+0.89)}}&94.45$\pm$0.17 \\

\textbf{\texttt{Receding Hairline }}& 69.72$\pm$0.78 &74.58$\pm$0.21& 78.86$\pm$0.40& 84.95$\pm$0.49& \cellcolor{blue!15} \textbf{87.00$\pm$0.12} & \cellcolor{green!10}86.84$\pm$0.33 \textcolor{red}{\tiny{(--0.16)}}& 85.15$\pm$1.31\\

\textbf{\texttt{Wavy Hair }}&73.10$\pm$0.56& 74.53$\pm$0.17& 77.39$\pm$0.15& \cellcolor{green!10}79.89$\pm$0.71 &79.42$\pm$0.12 & \cellcolor{blue!15}\textbf{80.80$\pm$1.26} \textcolor{red}{\tiny{(+1.38)}}&79.65$\pm$0.63 \\

\textbf{\texttt{Brown Hair}} &78.07$\pm$0.87 &78.93$\pm$1.24& 83.07$\pm$0.07& 83.83$\pm$0.66 &\cellcolor{green!10}85.30$\pm$0.47& \cellcolor{blue!15}\textbf{86.20$\pm$0.67} \textcolor{red}{\tiny{(+0.90)}}&84.87$\pm$0.07 \\ \hline
\end{tabular}
\vskip0.2cm 
\centering
\hskip0.05cm
\begin{tabular}{|
>{\centering\arraybackslash}p{32mm}|
>{\centering\arraybackslash}p{14mm}
>{\centering\arraybackslash}p{14mm}
>{\centering\arraybackslash}p{14mm}
>{\centering\arraybackslash}p{14mm}
>{\centering\arraybackslash}p{14mm}
>{\centering\arraybackslash}p{22mm}|
>{\centering\arraybackslash}p{14mm}|} 
\multicolumn{7}{c}{\textbf{(b) Worst-Group Accuracy (\%) }} \\ \hline
& \multicolumn{6}{c}{\cellcolor{gray!25}\textbf{Unsupervised}} & {\cellcolor{gray!25}\textbf{Supervised}}  \\ \hline
\textbf{Target} & \textbf{Baseline} & \textbf{LfF} & \textbf{George} &  \textbf{BPA} & \textbf{CFix} & \textbf{Ours} & \textbf{GDRO} \\ \hline
\textbf{\texttt{Double Chin}} &21.33$\pm$2.24 &28.24$\pm$0.46 &50.00$\pm$0.41 &67.78$\pm$0.91 &\cellcolor{green!10}74.26$\pm$3.94 &\cellcolor{blue!15}\textbf{78.69$\pm$0.32} \textcolor{red}{\tiny{(+4.43)}} & 72.94$\pm$1.14 \\
\textbf{\texttt{Pale Skin}}  &36.64$\pm$3.53 &43.26$\pm$1.40& 62.03$\pm$16.50& \cellcolor{green!10}88.60$\pm$1.48& 87.01$\pm$1.46& \cellcolor{blue!15}\textbf{89.76$\pm$0.62} \textcolor{red}{\tiny{(+1.16)}}& 87.68$\pm$2.37 \\
\textbf{\texttt{Wearing Necklace}}  &02.72$\pm$0.83& 06.67$\pm$2.07& 13.82$\pm$0.41& 41.93$\pm$2.47& \cellcolor{green!10}55.56$\pm$0.38& \cellcolor{blue!15}\textbf{60.98$\pm$0.20} \textcolor{red}{\tiny{(+5.42)}}& 24.34$\pm$7.81 \\
\textbf{\texttt{Wearing Hat}} & 85.12$\pm$0.31& 88.31$\pm$0.12& 92.93$\pm$0.76& 94.94$\pm$0.19&\cellcolor{green!10} 96.58$\pm$0.63& \cellcolor{blue!15}\textbf{96.75$\pm$0.33} \textcolor{red}{\tiny{(+0.17)}}&94.67$\pm$0.41 \\
\textbf{\texttt{Big Lips}} & 30.85$\pm$0.62 &38.54$\pm$0.18& 44.51$\pm$0.83& 56.99$\pm$3.05& \cellcolor{green!10} {57.27$\pm$0.58}&\cellcolor{blue!15} \textbf{57.79$\pm$0.15} \textcolor{red}{\tiny{(+0.52)}}& 47.55$\pm$1.03 \\
\textbf{\texttt{Bangs}} & 76.91$\pm$3.27& 82.37$\pm$0.52 &85.90$\pm$0.24& 92.21$\pm$1.24& \cellcolor{blue!15}\textbf{93.01$\pm$0.36}&\cellcolor{green!10}{92.66$\pm$0.38} \textcolor{red}{\tiny{(--0.35)}}& 92.12$\pm$1.03 \\
\textbf{\texttt{Receding Hairline}} & 35.69$\pm$0.35 &45.53$\pm$0.55& 57.30$\pm$0.90 &79.11$\pm$1.91& \cellcolor{green!10}84.15$\pm$0.82&\cellcolor{blue!15}\textbf{84.68$\pm$0.19} \textcolor{red}{\tiny{(+0.53)}}& 79.12$\pm$2.11 \\
\textbf{\texttt{Wavy Hair}} & 38.01$\pm$0.85& 45.24$\pm$0.83& 53.17$\pm$0.43 &65.74$\pm$1.13 &\cellcolor{green!10}69.92$\pm$0.38&\cellcolor{blue!15}\textbf{~80.05$\pm$0.39} \textcolor{red}{\tiny{(+10.13)}}& 66.79$\pm$1.62 \\
\textbf{\texttt{Brown Hair}}  &59.58$\pm$2.55& 60.68$\pm$3.62& 73.20$\pm$0.88& 71.50$\pm$0.97&\cellcolor{green!10} 79.18$\pm$0.50& \cellcolor{blue!15}\textbf{88.10$\pm$0.28} \textcolor{red}{\tiny{(+8.92)}}&78.92$\pm$1.61 \\ \hline
\end{tabular}
\label{tab:1}
\end{table*}

\begin{table*}
\centering
\mycustomfont	
\caption{Performance evaluation of \modelname{} against others on Waterbirds dataset. Cells in  \textcolor{blue!50}{blue} and \textcolor{green!70}{green} represent the best and second-best results, respectively.  Improvement gain compared to the second-best method is given between brackets in red.}
\begin{tabular}{|
>{\centering\arraybackslash}p{9mm}
>{\centering\arraybackslash}p{9mm}
>{\centering\arraybackslash}p{9mm}
>{\centering\arraybackslash}p{9mm}
>{\centering\arraybackslash}p{16mm}|
>{\centering\arraybackslash}p{9mm}|
>{\centering\arraybackslash}p{9mm}
>{\centering\arraybackslash}p{9mm}
>{\centering\arraybackslash}p{9mm}
>{\centering\arraybackslash}p{9mm}
>{\centering\arraybackslash}p{16mm}|
>{\centering\arraybackslash}p{9mm}|} \hline
\multicolumn{6}{|c|}{\textbf{Unbiased Accuracy (\%)}} &  \multicolumn{6}{c|}{\textbf{Worst-Group Accuracy (\%)}}  \\ \cline{1-12}
\multicolumn{5}{|c}{\cellcolor{gray!25}\textbf{Unsupervised}} & {\cellcolor{gray!25}\textbf{Sup.}} & \multicolumn{5}{c}{\cellcolor{gray!25}\textbf{Unsupervised}} & {\cellcolor{gray!25}\textbf{Sup.}} \\ \hline
\textbf{Baseline} & \textbf{LfF} &  \textbf{BPA} & \textbf{CFix} & \textbf{Ours} & \textbf{GDRO}  &\textbf{Baseline} & \textbf{LfF} &  \textbf{BPA} & \textbf{CFix} & \textbf{Ours} & \textbf{GDRO} 
\\ \hline
 87.99 & 85.05& 88.44& \cellcolor{blue!15}\textbf{92.17}& \cellcolor{green!10}91.49 \textcolor{red}{\tiny{(--0.68)}} & 89.20 & 73.34 & 60.00 & 79.16& \cellcolor{green!10}86.61& \cellcolor{blue!15}\textbf{87.76} \textcolor{red} {\tiny{(+1.15)}}& 85.27 \\ \hline
\end{tabular}
\label{tab:3}
\end{table*}
\subsection{Evaluation Metrics}
We evaluate the accuracy for each combination of target and bias attribute values ($y$, $a$), reporting the results as average-group accuracy (unbiased accuracy) and worst-group accuracy~\cite{capitani2024clusterfix,seo2022unsupervised}. The results are averaged over three independent runs.

\subsection{Baseline and Competitors}
\noindent \textbf{Baseline.} We compare \modelname{} against a vanilla-trained model, which does not incorporate any specific countermeasures for bias mitigation.

\noindent \textbf{Competitors.} We benchmark against several SOTA unsupervised debiasing methods: LfF~\cite{nam2020learning}, CFix~\cite{capitani2024clusterfix}, BPA~\cite{seo2022unsupervised}, and George~\cite{sohoni2020no}, where the latter three are clustering-based methods. Additionally, for an upper-bound performance comparison, we include GDRO~\cite{sagawa2019distributionally}, a method that optimizes worst-group performance over a distributionally robust uncertainty set using explicit bias supervision.

\subsection{Implementation Details} \label{imp_det}
{We use a ResNet18 model, pretrained on ImageNet, as the backbone for all methods. ResNet18 consists of four module layers, and we select the layers at the end of the second module and the final module as our shallow and deep layers, respectively. Before clustering, we apply PCA for dimensionality reduction. For preprocessing, images are resized to 224x224 for CelebA and Fitzpatrick, and 256x256 for Waterbirds, with standard augmentations including cropping, flipping, and normalization. We use official data splits and train \modelname{} with Adam (learning rate: $1\times10^{-4}$, batch size: 100, weight decay: 0.01) for 50 epochs. We set $\alpha$ to 0.1 and perform grid search for $\gamma$, yielding best results of 0.003-0.01 for the different labels in CelebA, 0.02 for Waterbirds, and 0.06 for Fitzpatrick.}

\begin{table}
\centering
\mycustomfont	
\caption{Performance evaluation of \modelname{} against others on Fitzpatrick dataset. Cells in  \textcolor{blue!50}{blue} and \textcolor{green!70}{green} represent the best and second-best results, respectively.  Improvement gain compared to the second-best method is given between brackets in red.}
\begin{tabular}{|
>{\centering\arraybackslash}p{6mm}
>{\centering\arraybackslash}p{6mm}
>{\centering\arraybackslash}p{6mm}|
>{\centering\arraybackslash}p{6mm}|
>{\centering\arraybackslash}p{6mm}
>{\centering\arraybackslash}p{6mm}
>{\centering\arraybackslash}p{6mm}|
>{\centering\arraybackslash}p{6mm}|} \hline
\multicolumn{4}{|c|}{\textbf{Unbiased Accuracy (\%)}} &  \multicolumn{4}{c|}{\textbf{Worst-Group Accuracy (\%)}}  \\ \hline
\multicolumn{3}{|c}{\cellcolor{gray!25}\textbf{Unsupervised}} & {\cellcolor{gray!25}\textbf{Sup.}} & \multicolumn{3}{c}{\cellcolor{gray!25}\textbf{Unsupervised}} & {\cellcolor{gray!25}\textbf{Sup.}} \\ \hline
\textbf{Baseline} & \textbf{LfF} &  \textbf{Ours} & \textbf{GDRO}  &\textbf{Baseline} & \textbf{LfF} &  \textbf{Ours} & \textbf{GDRO} \\ \hline
\multirow{2}{*}{77.90} & \cellcolor{green!10} & \cellcolor{blue!15}\textbf{82.69} & \multirow{2}{*}{77.12} & \multirow{2}{*}{36.73}  & \multirow{2}{*}{42.17} & \cellcolor{blue!15}\textbf{59.18} & {\cellcolor{green!10}} \\ 
 &  \multirow{-2}{*}{\cellcolor{green!10}78.61} & \cellcolor{blue!15} \textcolor{red}{\tiny{(+4.08)}}&  &  & &  \cellcolor{blue!15} \textcolor{red}{\tiny{(+3.06)}} & \multirow{-2}{*}{\cellcolor{green!10}56.12}  \\ \hline
\end{tabular}
\label{tab:4}
\end{table}

\subsection{Main Results}
\noindent \textbf{Qualitative Results.}  Table~\ref{tab:1} (top) demonstrates that \modelname{} outperforms all competitors, including the supervised GDRO and the previous unsupervised CFix, in most of the classification tasks on  CelebA dataset; e.g., for the \texttt{Wearing Necklace} target, our method achieves the highest accuracy at 71.14\%, surpassing CFix at 68.99\% and GDRO at 62.89\%. Similarly, for \texttt{Pale Skin}, our method leads with 92.77\%, compared to CFix at 91.17\% and GDRO at 90.55\%. 
Moreover, the improvements in worst-group accuracy achieved by \modelname{} are notable, as detailed in Table~\ref{tab:1} (bottom). Our method demonstrates gains, over second best performing method, of approximately 10\% for \texttt{Wavy Hair}, 9\% for \texttt{Brown Hair}, 5\% for \texttt{Wearing Necklace}, and 4\% for \texttt{Double Chin}. This achievement is significant given that \modelname{} does not specifically target worst-group accuracy, unlike methods such as George and GDRO, which explicitly optimize for this metric. These improvements in worst-group accuracy are achieved without compromising the performance across other groups, thereby ensuring a balanced enhancement, as reflected in the unbiased accuracy.

Additionally, experiments on Waterbirds in Table~\ref{tab:3} confirm the effectiveness of our method even in a challenging  controlled environment. We observe improvements in the worst-group performance, with gains of 1.15\% and 2.49\% compared to CFix and GDRO, respectively. 

In the medical domain, \modelname{} demonstrates superior performance on the Fitzpatrick dataset (Table~\ref{tab:4}), achieving the highest accuracies of 82.69\% and 59.18\%, surpassing the second-best method by 4.08\% and 3.06\% in terms of unbiased and worst-group accuracy, respectively.

\begin{figure}
    \centering
    \includegraphics[width=1\linewidth]{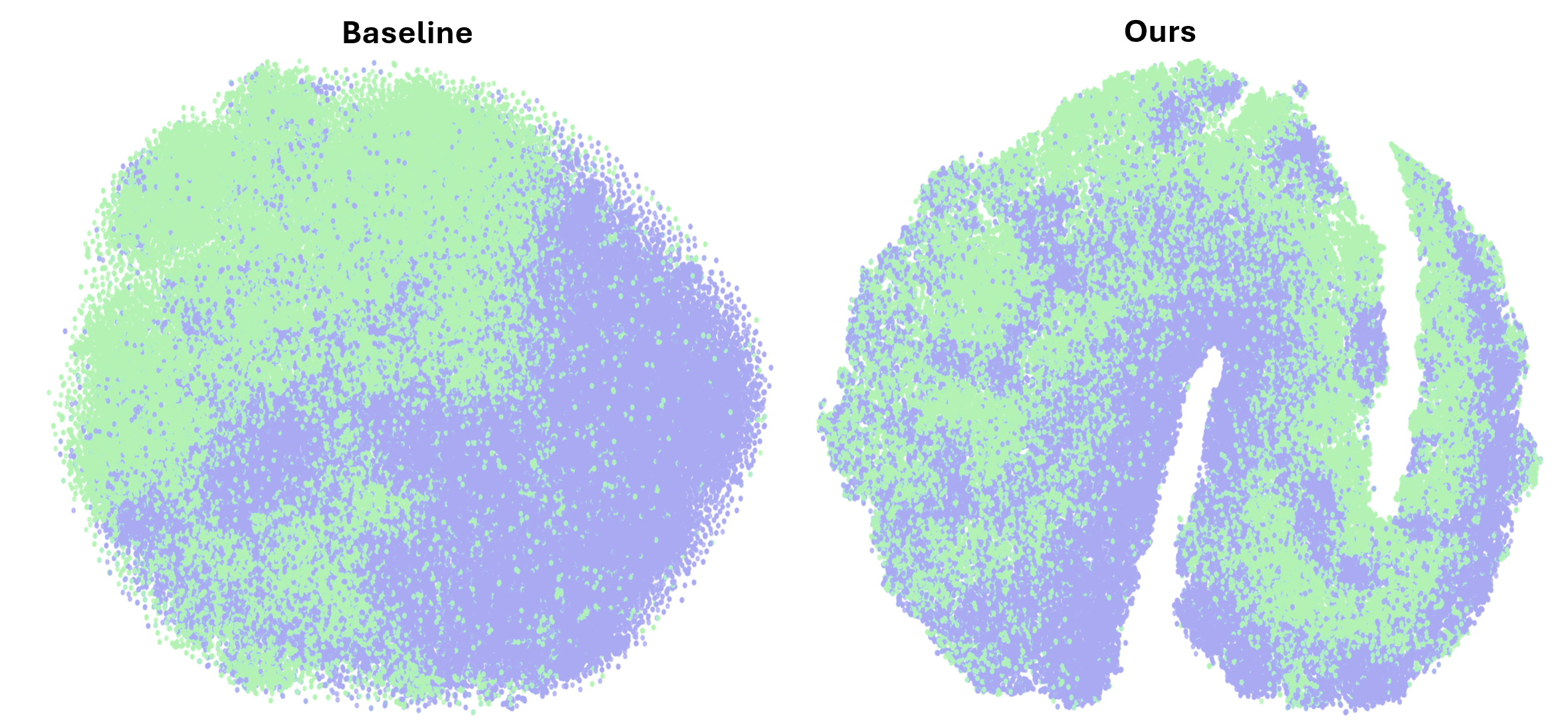}
    \caption{The t-SNE plots of feature embeddings for the baseline model (left) and our model (right) trained to classify \texttt{Blond Hair}. The plots display the distribution of samples with the target value \texttt{Blond Hair} = \texttt{False}. Blue and green colors represent female and male genders, respectively. Our \modelname{} promotes a better mix of samples with the same target but different bias attribute values, which reduces the bias.}
    \label{fig:tsne}
\end{figure}
\begin{figure*}
    \centering
    \includegraphics[width=\linewidth]{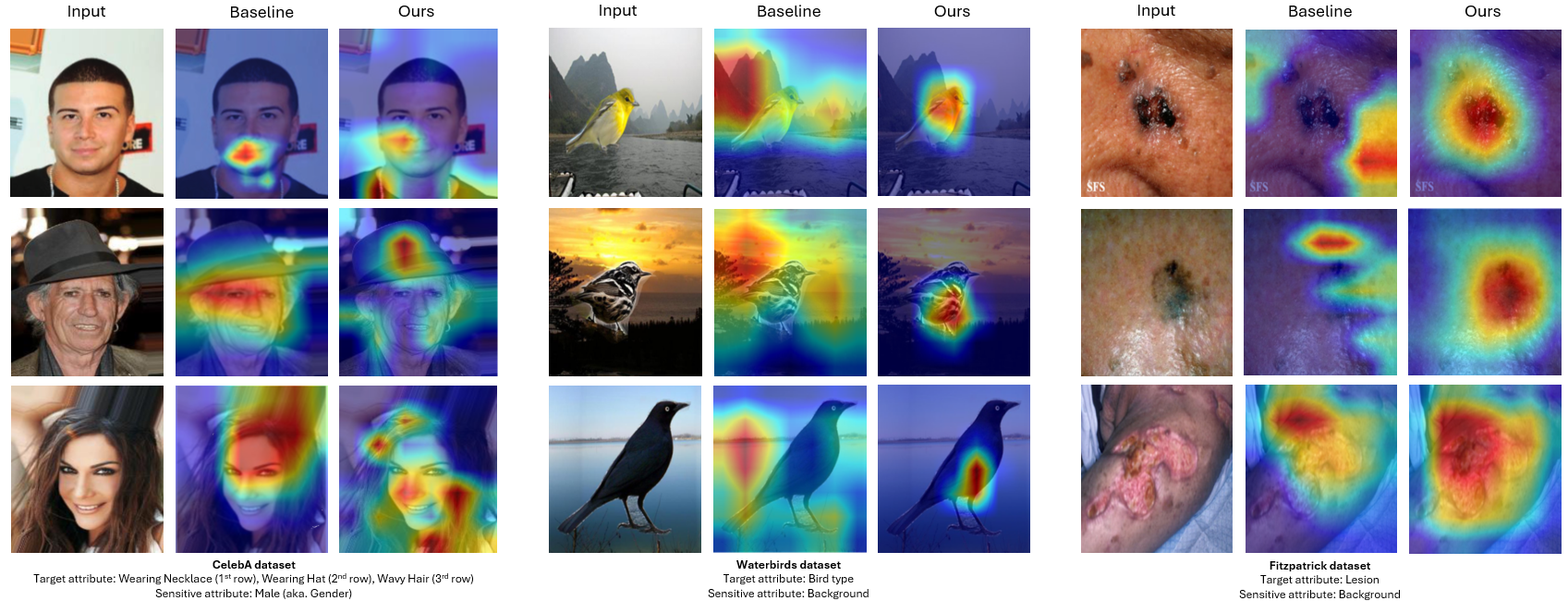}
    \caption{Visualization of the class activation maps generated by GradCAM for the baseline and \modelname{} (ours) on images from the CelebA (left), Waterbirds (middle), and Fitzpatrick (right) datasets.}
    \label{fig:gradcam_celeba}
\end{figure*}
\begin{table*}
\centering
\mycustomfont	
\caption{Performance evaluation (Unbiased Accuracy \%) of \modelname{} against others for \texttt{Blond Hair} classification with multiple bias attributes in the CelebA dataset. 
Cells in  \textcolor{blue!50}{blue} and \textcolor{green!70}{green} represent the best and second-best results, respectively. Improvement gain compared to the second-best method is given between brackets in red. }
\begin{tabular}{|
>{\centering\arraybackslash}m{10mm}
>{\centering\arraybackslash}m{10mm}|
>{\centering\arraybackslash}m{20mm}
>{\centering\arraybackslash}m{20mm}
>{\centering\arraybackslash}m{20mm}
>{\centering\arraybackslash}m{20mm}
>{\centering\arraybackslash}m{20mm}
>{\centering\arraybackslash}m{20mm}|} \hline
\multicolumn{2}{|c|}{\diagbox{\textbf{Method}}{\textbf{Bias}}} &   \texttt{\textbf{Male, Young}}  &  \texttt{\textbf{Male, Big Nose}} &  \texttt{\textbf{Male, Smiling}} & \texttt{\textbf{Male, Heavy Makeup}} &  \texttt{\textbf{Male, Wearing Lipstick}} &  \texttt{\textbf{Male, Wearing Necklace}} \\ \hline
\cellcolor{gray!25}& \textbf{Baseline} & 78.39 & 81.18 & 79.75 & 83.64 & 80.34 & 79.25 \\
\cellcolor{gray!25}& \textbf{LfF} & 81.21 & 84.10 & 82.91 & \cellcolor{blue!15}\textbf{88.82} & 84.13 & 81.03 \\
\multirow{-3}{*}{\cellcolor{gray!25}\rotatebox{90}{\textbf{Unsup.}}}  & \textbf{Ours} & \cellcolor{blue!15}\textbf{~~~~~~~~~88.35} \textcolor{red}{\tiny{(+0.39)}} & \cellcolor{blue!15}\textbf{~~~~~~~~~91.24} \textcolor{red}{\tiny{(+0.41)}} & \cellcolor{green!10}~~~~~~~~~89.78 \textcolor{red}{\tiny{(--1.95)}} & \cellcolor{green!10}~~~~~~~~~88.71 \textcolor{red}{\tiny{(--0.11)}} & \cellcolor{blue!15}\textbf{~~~~~~~~~87.21} \textcolor{red}{\tiny{(+1.28)}} & \cellcolor{green!10}~~~~~~~~~90.96 \textcolor{red}{\tiny{(--1.30)}} \\ \hline
\cellcolor{gray!25}\textbf{Sup.} & \textbf{GDRO} & \cellcolor{green!10}87.96 & \cellcolor{green!10}90.83 & \cellcolor{blue!15}\textbf{91.73} & 81.09 & \cellcolor{green!10}85.93 & \cellcolor{blue!15}\textbf{92.26} \\ \hline
\end{tabular}
\label{tab:5}
\end{table*}

\noindent\textbf{Feature Space Visualization.} 
In Figure~\ref{fig:tsne}, we present the t-SNE visualizations of feature embeddings generated by the baseline model (left) and our proposed model (right) on the CelebA dataset, specifically for the task of classifying the \texttt{Blond Hair} label. The embeddings are extracted from the penultimate layer of each model. In these plots, we focus solely on the negative examples (\texttt{Blond Hair} = \texttt{False}) to provide a clearer visualization of the data distribution. The colors represent the gender attribute, with blue indicating female and green indicating male. The baseline model's embeddings show a more segregated distribution based on gender, while our model's embeddings exhibit a more intermixed distribution of female and male samples within the same class. This intermixing indicates that \modelname{} is effective at reducing the influence of the gender bias attribute, promoting a more debiased representation of the data.

\noindent\textbf{Model Explainability.} 
In Figure~\ref{fig:gradcam_celeba}, we use GradCAM~\cite{selvaraju2020grad} to explore the interpretability of \modelname{} for the \texttt{Wearing Necklace}, \texttt{Wearing Hat}, and \texttt{Wavy Hair} classification tasks in CelebA dataset. Our method consistently focuses on regions highly relevant to the target attribute, while the baseline often emphasizes unrelated bias-based features. For instance, in \texttt{Wearing Necklace} classification task, our model focuses on the necklace, whereas the baseline focuses on the mouth. We further demonstrate the effectiveness of our method on images from the worst-group category in the Waterbirds and Fitzpatrick datasets (Figure~\ref{fig:gradcam_celeba}). In Waterbirds, our method clearly focuses on the bird, while the baseline emphasizes the background. Similarly, in Fitzpatrick, our method targets the specific skin lesion, whereas the baseline highlights the surrounding skin surface.

\noindent \textbf{Multiple Bias Attributes.} 
Thanks to the unsupervised design of \modelname{}, we can seamlessly evaluate its performance in multi-bias scenarios without modifying the network or training framework. Table~\ref{tab:5} shows the unbiased accuracy of \modelname{} compared to other methods for the \texttt{Blond Hair} classification task when multiple bias attributes are present. Our method consistently achieves superior results across various bias sets compared to other unsupervised methods. Additionally, it performs close to the supervised GDRO, which is very sensitive to the bias sets, as it trains a separate model for each bias. In contrast, \modelname{} can be applied to any bias set without further fine-tuning.

\begin{table*}
\centering
\mycustomfont	
\caption{Performance evaluation, given as unbiased (worst-group) accuracy, of \modelname{} from different ablation studies on CelebA, Waterbirds and Fitzpatrick datasets. } 
\begin{tabular}{|
>{\centering\arraybackslash}m{18mm}|
>{\centering\arraybackslash}m{33.2mm}|
>{\centering\arraybackslash}m{34mm}|
>{\centering\arraybackslash}m{33.2mm}|
>{\centering\arraybackslash}m{33.2mm}|} \hline
& \textbf{CelebA} (\texttt{\textbf{Double Chin}}) &  \textbf{CelebA } (\texttt{\textbf{Wearing Necklace}}) & \textbf{Waterbirds} & \textbf{Fitzpatrick} \\ \hline
\cellcolor{gray!25}\backslashbox{\textbf{Exp.}}{\textbf{Default}} &\cellcolor{gray!25} 86.19 (78.69) &\cellcolor{gray!25} 71.14 (60.98)  &\cellcolor{gray!25} 91.49 (87.76) &\cellcolor{gray!25} 82.69 (59.18) \\ \hline 
\textbf{\boldsymbol$\mathcal{A}$} & 84.37 (80.15) & 71.34 (61.87)  & 91.16 (88.32) & 81.36 (54.64) \\ \hline \hline 
\textbf{\boldsymbol$\mathcal{B}$} & 84.26 (78.13) & 68.36 (61.75)  & 90.26 (85.36) & 81.24 (57.83) \\  
\textbf{\boldsymbol$\mathcal{C}$} & 84.37 (77.62) & 69.22 (59.17)  & 91.67 (86.42) & 80.65 (58.27) \\  
\textbf{\boldsymbol$\mathcal{D}$} & 83.82 (78.45) & 68.29 (59.54)  & 89.28 (86.19) & 81.44 (58.67) \\  
\textbf{\boldsymbol$\mathcal{E}$} & 85.11 (80.62) & 71.88 (62.30)  & 90.85 (87.44) & 83.09 (60.31) \\  \hline \hline
\textbf{\boldsymbol$\mathcal{F}$} & 83.82 (69.83) & 70.32 (56.28)  & 88.65 (83.21) & 81.67 (57.24) \\ 
\textbf{\boldsymbol$\mathcal{G}$} & 84.61 (73.29) & 69.88 (58.39)  & 89.27 (84.69) & 82.35 (58.61) \\ \hline \hline
\textbf{\boldsymbol$\mathcal{H}$} & 83.82 (69.85) & 70.93 (56.36)  & 89.54 (83.29) & 80.35 (46.54) \\  
\textbf{\boldsymbol$\mathcal{I}$} & 83.45 (78.68) & 70.32 (59.11)  & 90.73 (84.68) & 82.11 (57.29) \\  \hline \hline
\textbf{\boldsymbol$\mathcal{J}$} & 84.00 (72.06) & 68.74 (56.37)  & 90.23 (85.41) & 81.68 (58.67) \\  
\textbf{\boldsymbol$\mathcal{L}$} & 83.63 (73.53) & 69.35 (59.21)  & 91.16 (86.32) & 82.35 (59.27) \\  
\textbf{\boldsymbol$\mathcal{M}$} & 85.17 (83.09) & 69.21 (63.48)  & 91.25 (86.27) & 82.24 (58.91) \\  \hline
\end{tabular}
\label{tab:6}
\end{table*}

\noindent \textbf{Linear Decodability.} As discussed in Sec~\ref{prelim_results}, we repeat the linear decodability experiments using our \modelname{}. Figure~\ref{fig:linear_dec} shows that bias decodability, whether for a single or combined attributes, is consistently lower with our method (orange bars) compared to the baseline (blue bars), demonstrating its effectiveness in mitigating bias across all layers. 

\subsection{Ablation Studies} \label{sec:ablations} 

We perform five sensitivity analyses on the CelebA (\texttt{Double Chin} and \texttt{Wearing Necklace}), Waterbirds, and Fitzpatrick datasets, summarized in Table~\ref{tab:6}.

\noindent \textbf{1. Cluster Assignment.} In Exp.~$\mathcal{A}$, we evaluate the impact of using true bias distributions versus pseudo clusters derived from the shallow layer on the performance of \modelname{}. Specifically, we replace the attribute-conditioned clusters $P_{a_k,y}$ in Eq.~\ref{eq:akd} with the true bias distributions, assuming they are available during training. The results demonstrate that utilizing our pseudo clusters derived from the shallow layer is generally effective, with performance comparable to or marginally better than using true bias distributions in Exp.~$\mathcal{A}$. Particularly notable improvements are observed in the Fitzpatrick dataset, where \modelname{} achieves a higher accuracy, especially in worst-group performance, compared to using true bias distributions. A possible explanation for this result lies in the inherent noise or imperfections in the true bias labels. Clustering through \modelname{} allows for a more nuanced understanding and adaptation to subtle variations within the dataset that may not be fully captured by the explicit bias labels, leading to improved performance.

\noindent \textbf{2. Distillation Depth.} 
We assess how different distillation depths impact \modelname{} by fixing the deep layer at the default position (layer 4) and systematically adjusting the shallow layer involved in the distillation process. Specifically, we evaluate the following shallow layer configurations: Layer 1 (Exp.~$\mathcal{B}$), layer 3 (Exp.~$\mathcal{C}$), the combination of layers 1 and 2 (Exp.~$\mathcal{D}$), and the combination of layers 1, 2, and 3 (Exp.~$\mathcal{E}$). Our findings are as follows: Interestingly, our method is effective across different shallow layer configurations; i.e., distilling knowledge from the deep layer into any shallow layer consistently improves debiasing, leading to superior worst-group accuracy compared to all SOTA methods across all datasets (Tables~\ref{tab:1},~\ref{tab:3},~\ref{tab:4}). This demonstrates the strong impact of our proposed deep-to-shallow distillation mechanism. 2) The multi-layer distillation in Exp.~$\mathcal{E}$ achieves the highest worst-group accuracy across the other configurations, which is expected as it guides all layers to learn debiased representations. However, it requires longer training and increased complexity due to the multi-layer clustering and distillation. 3) Based on average results across datasets, the default \modelname{} configuration (i.e., layer 2) offers the best performance balance while avoiding the computational overhead of Exp.~$\mathcal{E}$.

\noindent \textbf{3. Distance Metric Selection.} To assess the sensitivity of \modelname{} to the choice of distance metric used for the learning of debiased representations, we replace MMD in Eq.~\ref{eq:akd} with Kullback-Leibler (KL) divergence (Exp.~$\mathcal{F}$) and Mahalanobis distance (Exp.~$\mathcal{G}$). We observe that while both alternative metrics show a slight decline in unbiased accuracy, the use of Mahalanobis distance (Exp.$\mathcal{G}$) results in a relatively smaller decrease in worst-group accuracy compared to KL divergence (Exp.~$\mathcal{F}$). Overall, the results suggest that MMD is the most effective distance metric for our method, providing the best balance between unbiased accuracy and worst-group performance.

\noindent \textbf{4. Hybrid Loss Components.} 
In Exp.~$\mathcal{H}$ and $\mathcal{I}$, we evaluate the impact of omitting different components of the hybrid loss $\mathcal{L}_{hybrid}$ (Eq.~\ref{eq:hybrid}). We notice that omitting $\mathcal{L}_{AKD}$ (Exp.~$\mathcal{H}$) significantly decreases worst-group accuracy across all datasets, underscoring its crucial role in learning debiased representations. Conversely, omitting $\mathcal{L}_{KL}$ (Exp.~$\mathcal{I}$) results in less performance degradation, as this component primarily aids in knowledge transfer between logits, which is less essential for developing robust, debiased features.

\noindent \textbf{5. Number of Clusters.}  We investigate the performance of \modelname{} by varying the threshold value $\gamma$ to obtain different number of clusters per class. Specifically, we experiment with $\mathcal{K} =$ 2, 4 and 16 in Exps.~$\mathcal{J}$, $\mathcal{L}$, and $\mathcal{M}$, respectively. Note that the default values of $\gamma$ reported in Sec.~\ref{imp_det} result in $\mathcal{K}$ = 8 for the  CelebA and Waterbirds datasets, and $\mathcal{K}$ = 2 for the Fitzpatrick dataset. We observe that using a higher number of clusters (\(\mathcal{K} = 16\) in Exp.~$\mathcal{M}$) generally improves the worst-group accuracy across the datasets. However, this comes at the cost of decreased unbiased accuracy and increased computational complexity and training time. Conversely, fewer clusters (\(\mathcal{K} = 2\) in Exp.~$\mathcal{J}$) result in a notable drop in performance in the CelebA dataset, particularly in worst-group accuracy, while having less impact on the Waterbirds and Fitzpatrick datasets. This suggests that a moderate number of clusters is sufficient to capture the necessary attribute-conditioned variations while maintaining the model's overall robustness and efficiency.

\section{Conclusions}
We present a robust unsupervised debiasing framework that leverages feature clustering and self-distillation. \modelname{} is based on the  observation that images with similar non-target attribute labels cluster prominently in shallow neural network layers. Building on this, we introduce a novel clustering-based method, \modelname{}, which leverages these attribute-conditioned clusters to learn debiased representations using a self-distillation technique. Our technique enforces the distributions of these clusters to converge towards each other while simultaneously aligning them with the distribution of their respective class in the deepest layer, where more complex and predictive features reside. We demonstrate \modelname{}'s effectiveness through extensive experiments, outperforming previous debiasing methods, particularly in worst-group accuracy.

\noindent \textbf{Future Work.} While \modelname{} has proven robust across different shallow layer configurations, future work could develop a systematic approach or trainable module to automatically select the optimal layer(s) for enhanced debiasing, improving adaptability across a wider range of architectures and datasets.

%%%%%%%%% REFERENCES
{\small
\bibliographystyle{ieee_fullname}
\bibliography{egbib}
}

\end{document}